\documentclass[sigconf]{acmart}

\AtBeginDocument{%
  \providecommand\BibTeX{{%
    \normalfont B\kern-0.5em{\scshape i\kern-0.25em b}\kern-0.8em\TeX}}}

\setcopyright{acmcopyright}
\copyrightyear{2021}
\acmYear{2021}
\setcopyright{acmlicensed}\acmConference[MM '21]{Proceedings of the 29th
ACM International Conference on Multimedia}{October 20--24, 2021}{Virtual
Event, China}
\acmBooktitle{Proceedings of the 29th ACM International Conference on
Multimedia (MM '21), October 20--24, 2021, Virtual Event, China}
\acmPrice{15.00}
\acmDOI{10.1145/3474085.3475215}
\acmISBN{978-1-4503-8651-7/21/10}

\usepackage{balance}
\usepackage{epsfig}
\usepackage{graphicx}
\usepackage{amsmath}
\usepackage[american]{babel}
\usepackage{microtype}
\usepackage{xcolor}
\usepackage{color, colortbl}
\usepackage{multirow}

\usepackage{bm}
\usepackage{amsmath}
\usepackage{chngcntr}

\usepackage{paralist}
\usepackage[font=small]{caption}

\definecolor{cgreen}{RGB}{11,135,28}
\usepackage{algorithm,algpseudocode}
\usepackage{amsmath}
\usepackage{amsfonts}
 
\newcommand{\ie}{i.e.}
\newcommand{\eg}{e.g.}
\newcommand{\myparagraph}[1]{\vspace{3pt}\noindent{\bf #1}}

\newcommand{\jimmyadd}[1]{\textcolor{black}{ #1}}
\newcommand{\jimmyaddt}[1]{\textcolor{black}{#1}} 

\newcommand{\abcaddd}[1]{\textcolor{black}{#1}}
\newcommand{\abcaddpre}[1]{\textcolor{black}{#1}}

\newcommand{\wqz}[1]{\textcolor{black}{#1}}

\newcommand{\wenjiaadd}[1]{\textcolor{black}{ #1}}
\newcommand{\wenjia}[1]{\textcolor{cgreen}{[Wenjia: #1]}}

\newcommand{\CUT}[1]{}

\begin{document}
\fancyhead{}
\title{Group-based Distinctive Image Captioning \\ with  Memory Attention}


\author{Jiuniu Wang$^{1,2,3}$, Wenjia Xu$^{2,3}$, Qingzhong Wang$^{1,4}$, Antoni B. Chan$^{1,}$} 
\authornote{Corresponding author.}
\affiliation{\institution{$^1$ Department of Computer Science, City University of Hong Kong}}
\affiliation{\institution{$^{2}$ Aerospace Information Research Institute, Chinese Academy of Sciences}}
\affiliation{\institution{$^{3}$ University of Chinese Academy of Sciences}}
\affiliation{\institution{$^{4}$ Baidu Research}}
\affiliation{\institution{\{jiuniwang2-c,qingzwang2-c\}@my.cityu.edu.hk, xuwenjia16@mails.ucas.ac.cn, abchan@cityu.edu.hk}}

\begin{abstract}
Describing images using natural language is widely known as image captioning, which has made consistent progress due to the development of computer vision and natural language generation techniques. Though conventional captioning models achieve high accuracy based on popular metrics, \ie, BLEU, CIDEr, and SPICE, the ability of captions to distinguish the target image from other similar images is under-explored. To generate distinctive captions, a few pioneers employ contrastive learning or re-weighted the ground-truth captions, 
which focuses on one single input image.
However, the relationships between objects in a similar image group~(\eg, items or properties within the same album or fine-grained events) are neglected. In this paper, we improve the distinctiveness of image captions using a Group-based Distinctive Captioning Model~(\texttt{GdisCap}), which compares each image with other images in one similar group and highlights the uniqueness of each image. 
In particular, we propose a group-based memory attention~(\texttt{GMA}) module, which stores object features that are unique among the image group (\ie, with low similarity to objects in other images).
These unique object features are highlighted when generating captions, resulting in more distinctive captions.
Furthermore, the distinctive words in the ground-truth captions are selected to supervise the language decoder and \texttt{GMA}. Finally, we propose a new evaluation metric, distinctive word rate~(DisWordRate) to measure the distinctiveness of captions. Quantitative results indicate that the proposed method significantly improves the distinctiveness of several baseline models, and achieves the state-of-the-art performance on both accuracy and distinctiveness. Results of a user study agree with the quantitative evaluation and demonstrate the rationality of the new metric DisWordRate.
\end{abstract}



\maketitle

\section{Introduction}
\begin{figure}[tb]
	\begin{center}
		\includegraphics[width=\linewidth]{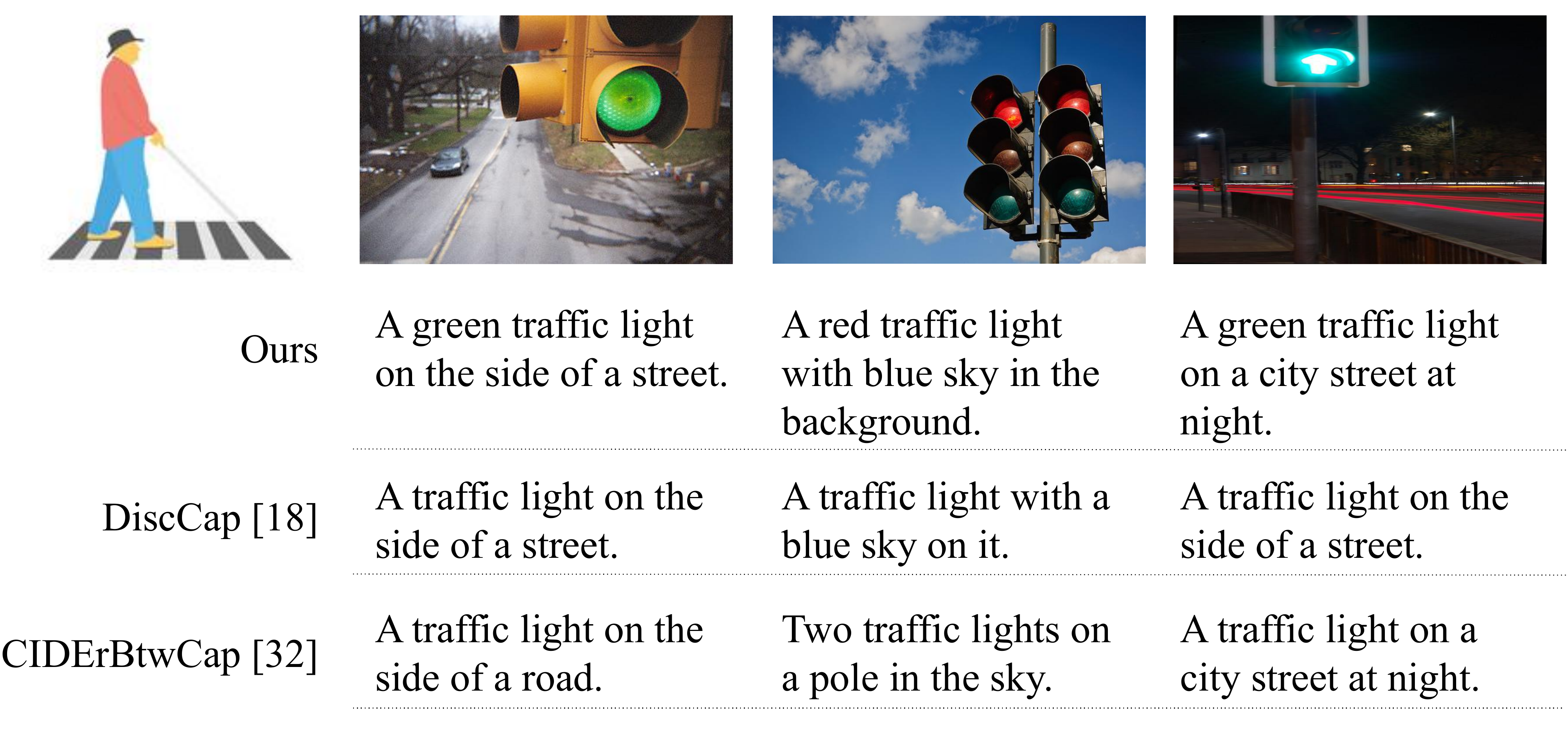}
	\end{center}
	\caption{Our model generates distinctive captions that can distinguish the target image from other similar images. Compared to current distinctive image captioning models \cite{2_disccap,17_wang2020compare}, our captions can specify the important details, e.g., the color and the environment of the traffic light, which can help a visually-impaired person to cross the street.}
	\label{fig:teaser_figure}
\end{figure}

The task of image captioning has drawn much attention from both the computer vision and natural language generation communities, and consistent progress has been made due to the development of vision and language techniques~\cite{NIC,spatt,updown,mrnn}. The fluent and accurate descriptions have aided many applications such as summarizing photo albums, tagging images on the internet, etc. An exciting application has been adopted on mobile phones, which speaks out what the cellphone camera sees\footnote{https://www.apple.com/accessibility/vision/}, to act as a tool describing the world for visually-impaired people. However, as pointed out in~\cite{17_wang2020compare,19_wang2019describing,18_luo2020analysis}, traditional captioning models that optimize the cross-entropy loss or reinforcement reward may lead to over-generic image captions.
Although the automatic image caption generators can accurately describe the image, they generate generic captions for images with similar semantic meaning, lacking the intrinsic human ability to describe the unique details of images \jimmyadd{to distinguish these images from others}. For instance, as shown in Figure~\ref{fig:teaser_figure}, simply mentioning the traffic light without explaining the specific meaning~(\eg, the color of the traffic light) cannot help visually-impaired people  to make a decision whether or not to cross the street. We argue that a model that describes the distinctive contents of each image is more likely to highlight the truly useful information. In this paper, we consider the goal of endowing image captioning models to generate distinctive captions. Here we refer to the {\em distinctiveness} of a caption as its ability to {\em describe the unique objects or context of the target image so as to distinguish it from other semantically similar images.}

Most of the existing image captioning models aim to generate a caption that best describes the semantics of a target image. Distinctiveness, on the other hand, requires
the caption to best match the target image among similar images, \ie, describing the distinctive parts of the target image. Some efforts have been made to generate {\em diverse} captions to enrich the concepts by employing conditional GAN~\cite{cgan1,3_cgan}, VAE~\cite{jain2017creativity,wang2017diverse} or reinforcement learning \cite{wang2019towards,wang2020diversity}. 
However, improving the diversity cannot guarantee distinctiveness, since rephrasing the expressions or enriching the vocabulary do not necessarily introduce novel and distinctive information. 
Several methods proposed to improve the distinctiveness by contrastive learning~\cite{2_disccap,1_contrastive,12_li2020context}, 
where \wenjiaadd{they either aggregate the contrastive image features with target image feature, or apply contrastive loss to suppress the estimated conditional probabilities of mismatched image-caption pairs~\cite{1_contrastive,2_disccap}.} 
However, the distractors are \wenjiaadd{either a group of images with scene graph that partially overlaps with the target image~\cite{12_li2020context}, or randomly selected unmatched image-caption pairs~\cite{1_contrastive,2_disccap}}, which are easy to distinguish.
~\cite{4_self-retrieval,9_PSST} introduce self-retrieval reward with contrastive loss, which requires the generated captions to retrieve the target image in a visual-language space. However, weighing too much on image retrieval could lead a model to repeat the distinctive words~\cite{17_wang2020compare}, which may hurt caption quality.

In this work, we consider the hard negatives, \ie, similar images that generally share similar semantics with the target image, and push the captions to clearly show the difference between these images. For instance, as shown in Figure~\ref{fig:teaser_figure}, the generated captions should specify the different aspects of the target image (\eg, different light colors) compared with other images that share similar semantics (\eg, images of traffic lights). 
Recently, the developments of transformer-based model and attention mechanism improve the accuracy of image captions. In this paper, we propose a plug-and-play distinctive memory attention module to extend the transformer-based captioning models,  
where we put high attention on distinctive objects detected in the target image -- object features in the target image with low similarity to object features of the similar images are considered more distinctive, and thus receive higher attention.

In summary, the contributions of this paper are three-fold: 
\begin{compactenum}
	\item We propose a Group-based Distinctive Captioning Model (\texttt{GdisCap}), which builds memory vectors from object regions that are weighted by distinctiveness among the image group, and then generates distinctive captions for the images in the group. 
	\item
	To enforce the weighted memory to contain distinctive object information, we further propose \jimmyadd{two distinctive losses}, where the supervision is the distinctive words occurring in the ground-truth (GT) captions.
	\item We conduct extensive experiments and user studies, demonstrating that the proposed model is able to generate distinctive captions. \wqz{In addition, our model also highlights the unique regions of each image, which is more interpretable.} 
\end{compactenum}


\section{Related work}
\myparagraph{Image captioning} bridges two domains---images and texts. Classical approaches usually extract image representations using Convolutional Neural Network~(CNN), then feed them into Recurrent Neural Network~(RNN) and output sequences of words~\cite{NIC,mrnn,karpathy2015deep}. Recent advances mainly focus on improving the image encoder and the language decoder. For instance,~\citet{updown} proposes bottom-up features, which are extracted by a pre-trained Faster-RCNN \cite{renNIPS15fasterrcnn} and a top-down attention LSTM, where an object is attended in each step 
when predicting captions. Apart from using RNNs as the language decoder,~\citet{aneja2018convolutional,wang2018cnn+,wang2018gated} utilize CNNs since LSTMs cannot be trained in a parallel manner. \citet{li2019entangled,cornia2020meshed} adopt transformer-based networks with multi-head attention to generate captions, which mitigate the long-term dependency problem in LSTMs and significantly improves the performance of image captioning. Recent advances usually optimize the network with a standard training procedure, where they pre-train the model with word-level cross-entropy loss~(XE) and fine-tune with reinforcement learning~(RL). However, as pointed out in~\cite{17_wang2020compare,1_contrastive,3_cgan}, training with XE and RL may encourage the model to predict an “average” caption that is close to all ground-truth (GT) captions, thus resulting in over-generic captions that lacks distinctiveness.

More relevant to our work are the recent works on group-based image captioning~\cite{12_li2020context,13_vedantam2017context,11_chen2018groupcap}, where a group of images is utilized as context when generating captions. \citet{13_vedantam2017context} generates sentences that describe an image in the context of other images from closely related categories. \citet{11_chen2018groupcap} summarizes the unique information of the target images contrasting to other reference images, and~\citet{12_li2020context} emphasizes both the relevance and diversity. Our work is different in the sense that we simultaneously generate captions for each image in a similar group, and highlight the difference among them by focusing on the distinctive image regions. \wenjiaadd{Both \citet{11_chen2018groupcap,13_vedantam2017context} extract one image feature from the FC layer for each image, where all the semantics and objects are mixed up. While our model focuses on the object-level features and explicitly finds the unique objects that share less similarity with the context images, leading to fine-grained and concrete distinctiveness.} 

\myparagraph{Distinctive image captioning} aims to overcome the problem of generic image captioning, by describing sufficient details of the target image to distinguish it from other images.
\citet{1_contrastive} promotes the distinctiveness of image caption by contrastive learning.
The model is trained to give a high probability to the GT image-caption pair and lower the probability of the randomly sampled negative pair.
\citet{2_disccap,4_self-retrieval,9_PSST} takes the same idea that the generated caption should be similar to the target image rather than other distractor images in a batch, and applies caption-image retrieval to optimize the contrastive loss. However, the distractor images are randomly sampled in a batch, which can be easily distinguished from the target images. In our work, we consider {\em hard negatives} that share similar semantics with the target image, and push the captions to contain more details and clearly show the difference between these images. 
More recently, \cite{17_wang2020compare} proposes to give higher weight to the distinctive GT captions during model training.
\citet{11_chen2018groupcap} 
models the diversity and relevance among positive and negative image pairs in a language model, with the help of a visual parsing tree \cite{2017StructCap}.
In contrast to these works, our work compares a group of images with a similar context, and highlights the unique object regions in each image to distinguish them from each other. 
That is, our model infers which object-level features in each image are unique among all images in the group.
\abcaddd{Our model is applicable to most of the transformer-based captioning models.}





\myparagraph{Attention mechanism} applies visual attention to different image regions when predicting words at each time step, which has been widely explored in image captioning \cite{spatt,you2016image,chen2017sca,15_guo2020normalized,pan2020x}. For instance, \citet{you2016image} adopts semantic attention to focus on the semantic attributes in image. \citet{updown} exploits object-level attention with bottom-up attention, then associates the output sequences with salient image regions via top-down mechanism. More recently, self-attention networks introduced by Transformers~\cite{vaswani2017attention} are widely adapted in both language and vision tasks~\cite{dosovitskiy2020image,ye2019cross,ramachandran2019stand,yang2020bert,su2019vl}. \citet{15_guo2020normalized} normalizes the self-attention module in the transformer to solve the internal covariate shift. \citet{huang2019attention} weights the attention information by a context-guided gate.
These works focus on learning self-attention between every word token or image region in one image. 
\citet{12_li2020context} migrates the idea of self-attention to visual features from different images, and averages the group \abcaddd{{\em image-level}} vectors with self-attention to detect prominent features. In contrast, in our work, we take a further step by learning memory attention among the R-CNN \abcaddd{{\em object-level}} features~\cite{updown} extracted from similar images, to highlight the prominent features that convey distinguishing semantics in the similar image group.

\begin{figure*}[t]
	\begin{center}
		\includegraphics[width=\linewidth]{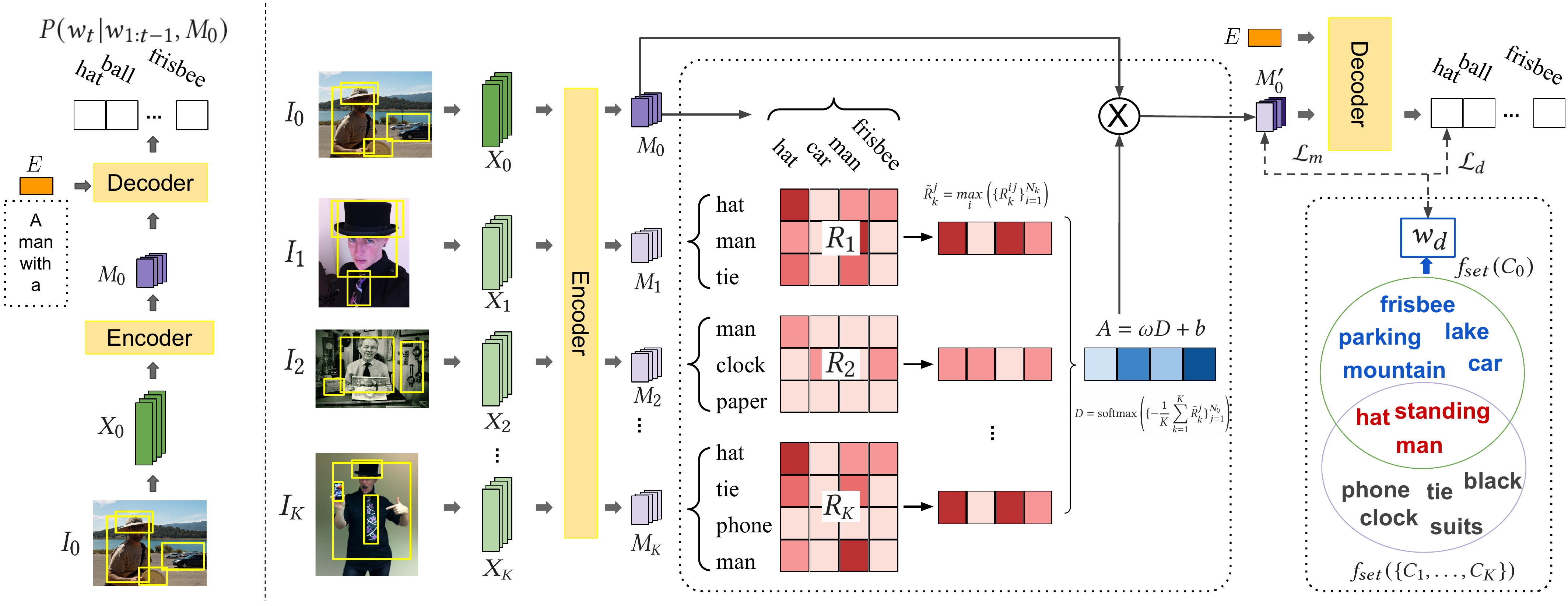}
	\end{center}
	\caption{Left: the standard transformer-based captioning model, where the target image features $X_0$ are the region-based visual features extracted via RoI pooling from Faster R-CNN. Right: our Group-based Distinctive Captioning Model (\texttt{GdisCap}), which consists of a group-based memory attention (\texttt{GMA}) module that weights the memory vectors according to their similarity with other similar images. \abcaddpre{Our model takes a group of images as input, and outputs one caption for each image. Only one target memory, one decoder and one output caption are shown here to reduce clutter.} 
	}
	\label{fig:model}
\end{figure*}

\section{Methodology}


We present the framework of our proposed Group-based Distinctive Captioning model (\texttt{GdisCap}) in Figure~\ref{fig:model}. Our model aims at generating distinctive captions for each image within a group of semantically similar images. Given an image group with $K+1$ images, denoted as $\{I_0, I_1, \dots, I_K \}$,  \texttt{GdisCap} generates distinctive captions for each image. 
Different from the conventional image captioning task, the generated captions should describe both the salient content in the target image, and also highlight the uniqueness of the target image (\eg, $I_0$) compared to other $K$ images (\eg, $I_1$ to $I_K$) in the same group. Specifically, during training, each image in the group is treated equally, and we use \abcaddd{each} image as a target. In Figure~\ref{fig:model}, we show an example where $I_0$ is the target image.


To achieve the goal of distinctive image captioning, we first construct similar image groups, then we employ the proposed Group-based Memory Attention~(\texttt{GMA})
to extract the distinctive object features. Finally, we design two distinctive losses to further encourage generating distinctive words.

\subsection{Similar image group}

Similar image group was first introduced in \cite{17_wang2020compare} to evaluate the distinctiveness of the image captions.
\abcaddd{For training, our model handles \jimmyadd{several} similar image groups as one batch, simultaneously using each image in the group as a target image.}
Here, we dynamically construct the similar image groups during training as follows:

(1) To construct a similar image group, we first randomly select one image as the target image $I_0$, and then retrieve its $K$ nearest images through a semantic similarity metric, measured by the visual-semantic retrieval model VSE++~\cite{faghri2017vse++}, as in \cite{17_wang2020compare}. In detail, given the target image $I_0$, we use VSE++ to retrieve those captions that well describe $I_0$ among all human-annotated captions, and then the corresponding images of those captions are the nearest images.

(2)
Due to the distribution inequality, the images sharing similar semantic meaning will form clusters in the VSE++ space. The images in the cluster center may be close to many other images, and to prevent them from dominating the training, the $K+1$ images that are used to create one similar image group are removed from the image pool, so they will not be selected when constructing other groups. In this way, each image will belong to only one group, \abcaddd{with no duplicate images appearing in one epoch.}\footnote{\jimmyadd{When almost all images are selected, the remaining images are not similar enough to construct groups. We regard them as target images one by one, and find similar images from the whole image pool.
}}

Each data split (training, validation, test) is divided into similar image groups independently.
For each training epoch, we generate new similar image groups to encourage training set diversity.

\subsection{Group-based Distinctive Captioning Model} 

Here we introduce the group-based distinctive captioning model (\texttt{GdisCap}), and how we incorporate the Group-based Memory Attention~(\texttt{GMA}) module that encourages the model to generate distinctive captions. Notably, the \texttt{GMA} module can serve as a plug-and-play tool for distinctive captioning, which can be applied to most existing transformer-based image captioning models. 

\subsubsection{Transformer-based Image Captioning}

Our captioning model is build on a transformer-based architecture~\cite{cornia2020meshed}, as illustrated in Figure~\ref{fig:model}~(left). The model can be divided into two parts: an image {\em Encoder} that processes input image features, and a caption {\em Decoder} that predicts the output caption word by word. In transformer-based architecture,  {\em Encoder} and {\em Decoder} are both composed of several multi-head attention and MLP layers. 

Here we take the bottom-up features~\cite{updown} extracted by Faster R-CNN \cite{renNIPS15fasterrcnn} as the input. Given an image $I$, let $X=\{x^i\}_{i=1}^N$ denotes the object features, where $N$ is the number of region proposals and $x^i \in R^{d}$ is the feature vector for the $i$-th proposal. The output of the $l$-th encoder layer is calculated as follows:
\begin{align}
	O^{att}_{l} &= \textbf{LN}\left(X_{l-1} + \textbf{MH}\left(\mathbf{W_q}X_{l-1}, \mathbf{W_k}X_{l-1}, \mathbf{W_v}X_{l-1}\right)\right), \\
	X_{l} &= \textbf{LN}\left(O^{att}_l + \textbf{MLP}\left(O^{att}_l\right)\right),
\end{align}
where $\textbf{LN}(\cdot)$ denotes layer normalization, $\textbf{MLP}(\cdot)$ denotes a multi-layer perceptron, and $\textbf{MH}(\cdot)$ represents the multi-head attention layer. $\mathbf{W_q}, \mathbf{W_k}, \mathbf{W_v}$ are learnable parameters.

The {\em Encoder} turns $X$ into memory vectors $M =\{m^i\}_{i=1}^N$, where 
$m^i \in R^{d_m}$ encodes the information from the $i$-th object proposal $x^i$, and is affected by other objects in the multi-head attention layers, which contains both single object features and the relationships among objects. 
According to the memory  $M$ and the embedding $E$ of the previous word sequence $\{w_1, \dots, w_{t-1} \}$, the {\em Decoder} generates the $v$-dimensional word probability vector $P_t = P(w_t|w_{1:t-1},M)$ 
at each time step $t$, where $v$ is the size of vocabulary. 

\subsubsection{Group-based Memory Attention}

The goal of the group-based memory attention is to highlight the distinctive features of the target image that do not appear in other similar images. For instance, in Figure~\ref{fig:model}~(right), the concept of \textit{man} and \textit{hat} that appear in the target image $I_0$ are also shared in other similar images, but \textit{frisbee} and \textit{cars} are unique for $I_0$ and can distinguish $I_0$ from other images. However, the standard captioning model in Figure~\ref{fig:model}~(left) cannot highlight those objects, since each memory vector $m_0^i$ for different image regions are treated equally when fed into the {\em Decoder}. 

In this work, we aim to give more attention to the distinctive image regions when generating captions.  Hence, the model will describe the distinctive aspects of the input image instead of only describing the most salient regions. 
To this end, we propose a Group-based Memory Attention~(\texttt{GMA}) module (see Figure~\ref{fig:model} (right)), where the attention weight for each object region is obtained by calculating the distinctiveness of its memory vector $m_0^i$. Then we encourage the model to generate distinctive words associated with the unique object regions. 
Specifically, the \texttt{GMA} produces 
distinctive attention $A = \{a_i\}_{i=1}^{N_0} \in R^{N_0}$ for memory vectors $M_0$. When generating captions, instead of using $M_0$, a weighted target memory is fed into the decoder:
\begin{align}
	M_0'=\{a_i \cdot m_0^i\}_{i=1}^{N_0}.
	\label{equ:weight}
\end{align}



\myparagraph{Computing distinctive attention.}
To compute the distinctive attention, we need to compare the objects in the target image with those in the similar images.
As shown in Figure~\ref{fig:model}~(right),  the target image $I_0$ and its similar images $\{I_k\}_{k=1}^K$ are transferred into memory vectors $M_0=\{m_0^j\}_{j=1}^{N_0}$ and 
$M_k=\{m_k^i\}_{i=1}^{N_k}$, $k=1...K$, via the image encoder, where 
$N_k$ denotes the number of objects in the $k$-th image. 
The \texttt{GMA} first measures the similarity \jimmyaddt{$R_k \in R^{N_k \times N_0}$}  of each target memory vector $m_0^j$ and each memory vector $m_k^i$ in similar images
via cosine similarity:
\begin{align}
	R^{ij}_k = \cos(m^i_k, m^j_0) \,,
	\label{eqn:R}
\end{align}
where $m_0^j \in R^{d_m}$ is the $j$-th vector in $M_0$ (\eg, memory vectors for \textit{hat}, \textit{car}, \textit{man} and \textit{frisbee} in Figure~\ref{fig:model}), and $m_k^i$ is the $i$-th vector in $M_k$ (\eg, memory vectors for \textit{hat}, \textit{man}, and \textit{tie} from $M_k$ in Figure~\ref{fig:model}). 
%

\wenjiaadd{The similarity matrix reflects how common an object is -- a common object that occurs in many images is not distinctive for the target image. 
	For example, as shown in Figure~\ref{fig:model} (right), \texttt{hat} is less distinctive since it occurs in multiple images, while \texttt{car} is an unique object that only appear in target image.
}
To summarize the similarity matrix, we compute an object-image similarity map $\tilde{R}_k \in R^{N_0}$ as
\begin{align}
	\tilde{R}_k^j = \jimmyaddt{\mathop{max}\limits_{i} \left ( \{{R_k^{ij}}\}_{i=1}^{N_k} \right)}\,,
\end{align} 
\abcaddpre{where $\tilde{R}_k^j$ is the similarity of the best matching object-region in image $I_k$ to region $j$ in the target image $I_0$.}

We assume that 
objects with higher similar scores are less distinctive. Hence, the distinctiveness scores $D \in R^{N_0}$ for each memory vector $m_0^i$ are computed by softmax of the average of object-image similarity maps:
\begin{align}
	\jimmyaddt{D = \mathrm{softmax}\left (\{-\frac{1}{K}\sum_{k=1}^{K}{\tilde{R}_k^j}\}_{j=1}^{N_0} \right )\,,}
\end{align}
where higher scores indicates higher distinctiveness, i.e., lower average similarity to other similar images.
\jimmyadd{Note that the values of $D$ are in range $(0,1)$ due to the softmax function.}
Finally, the distinctive attention $A$ for each target memory vector in (\ref{equ:weight}) is calculated as:
\begin{align}
	A&=\omega D + b \,,
\end{align}
where $\omega$ and $b$ are two learnable parameters. 
\abcaddpre{The bias term $b$ controls the minimum value of $A$, i.e., the base attention level for all regions, while $\omega$ controls the amount of attention increase due to the distinctiveness.} \jimmyadd{We clip $\omega$ and $b$ to be non-negative, so that the attention value in A is non-negative.}

\subsection{Loss functions}


Two typical loss functions for training image captioning are  cross-entropy loss and reinforcement loss. 
Since these two losses only use the GT captions of the image for supervision, they may encourage the generated captions to mimic GT captions, resulting in over-genericness, \ie, lack of distinctiveness. To address this issue, we take a step further to define distinctive words and explicitly encourage the model to learn more from these words. 
In this section, we first review the two typical loss functions used in captioning models, and then present our proposed Distinctive word loss (DisLoss) and Memory Classification loss (MemCls) for training our \texttt{GMA} module.


\myparagraph{Cross-Entropy Loss.} 
Given the $i$-th GT caption of image $I_0$, $C^i_0=\{w_t\}_{t=1}^T$,
the cross-entropy loss is
\begin{align}
	\mathcal{L}_{xe}=  - \sum\limits_{t = 1}^T \log  P(w_t|w_{1:t-1}, M_0')\,,
\end{align}
where $P(w_t|w_{1:t-1}, M_0)$ denotes the predicted probability of the word $w_t$ conditioning on the previous words $w_{1:t-1}$ and the weighted memory vectors $M_0'$, as generated by the caption {\em Decoder}.

\myparagraph{Reinforcement learning loss.}
Following~\cite{rennie2017self}, we apply reinforcement learning to further improve the accuracy of our trained network using the loss:
\begin{align}
	\mathcal{L}_{r} = -E_{\hat{c}\sim p(c|I)} \left[\frac{1}{d_c}\sum\limits_{i = 1}^{d_c} g(\hat{c},C_0^i)\right],
	\label{eqn:newreward}
\end{align}
where $g(\hat{c},C_0^i)$ is the CIDEr value between the predicted caption $\hat{c}$ and the $i$-th GT $C_0^i$, and $d_c$ denotes the number of GT captions.


\myparagraph{Distinctive word loss (DisLoss).} 
In this work, we focus on the distinctive words ${w_{d}}$ that appear in captions $C_0$ of target image but not in captions $\{C_1, \dots, C_K\}$ of similar images.  We define the distinctive word set as
\begin{align}
	w_{d} = f_{set}(C_0) - f_{set}(\{C_1, \dots, C_K\}) \,,
\end{align}
where $f_{set}(\cdot)$ denotes the function that converts the sentence into word set, \abcaddpre{and ``$-$'' means set subtraction.}

In the training phase, we explicitly encourage the model to predict the distinctive words in $w_{d}$ by optimize the distinctive loss ${\mathcal L}_{d}$, 
\begin{align}
	{\mathcal L}_{d} = - \sum\limits_{t = 1}^T \sum\limits_{i = 1}^u \log P(w_t = w_{d}^i | w_{1:t-1}, M_0')  \,,
\end{align}
\jimmyadd{where $w_{d}^i$ denotes the $i$-th distinctive word in $w_{d}$, and $P(w_t = w_{d}^i | w_{1:t-1}, M_0')$ denotes the probability of predicting word $w_{d}^i$ as the $t$-th word in sentence.} $u$ is the number of words in $w_{d}$, and $T$ is the length of the sentence.

\myparagraph{Memory classification loss (MemCls).}
\abcaddpre{In order to generate distinctive captions, the {\em Decoder} requires the \texttt{GMA} to produce memory contents containing distinctive concepts. 
	However, the supervision on the \texttt{GMA} through the {\em Decoder} could be too weak, which may allow the \texttt{GMA} to also produce non-useful information, \eg, highlighting too much background or focusing on small objects that are not mentioned in the GT captions.}
%
To improve the distinctive content produced by the \texttt{GMA}, we introduce an {\em auxiliary classification task} that predicts the distinctive words from the weighted memory vectors $M_0'$ of the \texttt{GMA}, 
\begin{align}
	P_M = f_{MC}(M_0') \,,
\end{align}
where $P_M$ denotes the word probability vector and $f_{MC}$ is the classifier.
To associate the memory vectors with distinctive words, we employ the multi-label classification loss ${\mathcal L}_{m}$ to train the classifier,
\begin{align}
	{\mathcal L}_{m} =  - \sum\limits_{i = 1}^u {\log (P_{M, w_{d}^i}) }\,,
\end{align}
where $P_{M, w_{d}^i}$ is the predicted probability of the $i$-th distinctive word.


\myparagraph{The final loss.} The final training loss ${\mathcal L}$ is formulated as 
\begin{align}
	{\mathcal L} = \alpha_{c}{\mathcal L}_{xe} + \alpha_{r}{\mathcal L}_{r} + \alpha_{d}{\mathcal L}_{d} + \alpha_{m}{\mathcal L}_{m} \,,
\end{align}
where $\{\alpha_{c}, \alpha_{r}, \alpha_{d}, \alpha_{m}\}$ are hyper-parameters for their respective losses.
The training procedure has two stages. In the first stage, we set $\alpha_{c} = 1$ and $\alpha_{r} =0$, so that the network is mainly trained by cross-entropy loss $\alpha_{c}$.  In the second stage, we set $\alpha_{c} = 0$ and $\alpha_{r} =1$, so that the parameters are mainly optimized by reinforcement learning loss ${\mathcal L}_{r}$.
We adaptively set $\{\alpha_d, \alpha_m\}$ so that
$\alpha_{d} {\mathcal L}_{d}$ and $\alpha_{m} {\mathcal L}_{m}$ are one quarter of ${\mathcal L}_{xe}$ (or ${\mathcal L}_{r}$).

\abcaddd{During training, each mini-batch comprises several similar image groups, with the loss aggregated over each image as a target in its group. Details for processing one image group are in supplemental.}

\section{Experiments}
In this section, we first introduce the implementation details and dataset preparation, then we quantitatively evaluate the effectiveness of our model by an ablation study and a comparison with other state-of-the-art models.


\subsection{Implementation details}
Following~\cite{updown}, we use the spatial image features extracted from Faster-RCNN~\cite{renNIPS15fasterrcnn} \jimmyadd{with dimension $d=2048$}. Each image usually contain around $50$ object region proposals, i.e. $N_k \approx 50$. \jimmyadd{Each object proposal has a corresponding memory vector with dimension $d_m = 512$. 
	\abcaddd{We set $K=5$ for constructing the similar image groups.} 
	The values in distinctive attention $A$ are mostly in the range of $(0.5, 0.9)$.} To verify the effectiveness of our model, we conduct experiments on four baseline methods~(\ie, Transformer~\cite{vaswani2017attention}, M$^2$Transformer~\cite{cornia2020meshed}, Transformer + SCST~\cite{vaswani2017attention} and M$^2$Transformer + SCST)~\cite{cornia2020meshed}. 
\jimmyadd{Our experimental settings (\eg, data preprocessing and vocabulary construction) follow these baseline models. We apply our \texttt{GMA} module on Transformer model and all three layers in M$^2$Transformer model}. Note that our model is applicable to most of the transformer-based image captioning models, and we choose these four models as the baseline due to their superior performance on accuracy-based metrics.

\subsection{Dataset and metrics}

\subsubsection{Dataset}
We conduct the experiments using the most popular dataset---MSCOCO, which contains $12,387$ images, and each image has $5$ human annotations. Following \cite{updown}, we split the dataset into $3$ sets---$5,000$ images for validation, $5,000$ images for testing and the rest for training.
\wenjiaadd{When constructing similar image groups for test set, we adopt the same group split as~\cite{17_wang2020compare} for a fair comparison.}

\subsubsection{Metrics}
We consider two groups of metrics for evaluation. 
The first group includes the metrics that evaluate the accuracy of the generated captions, such as CIDEr and BLEU.
The second group assesses the distinctiveness of captions. For the latter, CIDErBtw~\cite{17_wang2020compare} calculates the CIDEr value between generated captions and GT captions of its similar image group. However, CIDErBtw only works when comparing two methods with similar CIDEr value, \eg, a random caption that has lower overlap with the GT captions will be considered as distinctive since it achieves lower CIDErBtw. Hence, we propose two new distinctiveness metrics as follows. 

\myparagraph{CIDErRank.} A distinctive caption $ \hat{c}$ (generated from image $I_0$) should be similar to the target image's GT captions $C_0$, while different from the GT captions of other images in the same group $\{C_1, \dots, C_K\}$. Here we use CIDEr values $\{{s_k}\}_{k=0}^K$ to indicate the similarity of the caption $c$ with GT captions in images group as
\begin{align}
	s_k&=\frac{1}{d_c}\sum\limits_{i = 1}^{d_c} g(\hat{c},C_k^i)\,,
\end{align}
where $g(\hat{c},C_k^i)$ represents the CIDEr value of predicted caption $\hat{c}$ and $i$-th GT caption in $C_k$.
We use the rank of $s_0$ in $\{{s_k}\}_{k=0}^K$ to show the distinctiveness of the models as
\begin{align}
	r&=f_{rank} \left (s_0, \{{s_k}\}_{k=0}^K \right )\,,
\end{align}
where $f_{rank}(\cdot)$ means $s_0$ is the $r$-th largest value in $\{{s_k}\}_{k=0}^K$.
The best rank is 1, indicating the generated caption $\hat{c}$ is mostly similar to its GT captions and different from other captions, while the worst rank is $K+1$.
Thus, the average $r$ reflects the performance of captioning models, with more distinctive captions having lower CIDErRank.

\myparagraph{DisWordRate.} We design this metric based on the assumption that using distinctive words should indicate that the generated captions are distinctive. The {\em distinctive word rate} (DisWordRate) of a generated caption $\hat{c}$ is calculated as:
\begin{align}
	DisWordRate= \max_i \frac{{|w_d \cap \hat{c}\cap C_0^i|}}{{|w_d \cap C_0^i|}}, \quad i=1,\dots, d_c \,,
\end{align}
where $d_c$ is the number of sentence in $C_0$, $|w_d \cap \hat{c}|$ represents the number of elements in ${w_{d}}$ that appear in $\hat{c}$.
Thus, DisWordRate reflects the percentage of distinctive words in the generated captions.

\subsection{Main results}

\begin{table}[tb]
	\begin{center}
		\resizebox{240pt}{80pt}{
			\begin{tabular}{c|ccc|ccc}
				\textbf{Method} &\textbf{D(\%)$\uparrow$}& \textbf{CR$\downarrow$} & \textbf{CB$\downarrow$} & \textbf{C$\uparrow$}   &  \textbf{B3$\uparrow$} & \textbf{B4$\uparrow$} \\
				\hline
				Transformer~\cite{vaswani2017attention} & 16.8 & 2.47 & 74.8 & 111.7  & 45.1 & 34.0 \\
				\hspace{2mm}  + GdisCap (ours) & \textbf{19.5} & 2.42 & \textbf{70.9} & 107.3     & 43.4 & 32.7 \\
				\hline
				M$^2$Transformer~\cite{cornia2020meshed}* & 16.4  & 2.52 & 76.8 &  111.8   & 45.2 & 34.7 \\
				\hspace{2mm}  + GdisCap (ours)  & 18.7 & 2.43  & 72.5  & 109.8  & 45.6 & 34.7 \\
				\hline
				Transformer + SCST~\cite{vaswani2017attention} & 14.7 & 2.38 & 83.2 & 127.6 & \textbf{51.3} & \textbf{38.9} \\
				\hspace{2mm}  + GdisCap (ours) & 16.5 & 2.36 & 81.7 & 127.0 & 50.7 & 38.4 \\
				\hline
				M$^2$Transformer + SCST~\cite{cornia2020meshed}* & 17.3 & 2.38 & 82.9  & \textbf{128.9}  & 50.6 & 38.7 \\
				\hspace{2mm}  + GdisCap (ours)  & 18.5 & \textbf{2.31} & 81.3  & 127.5   & 50.0 & 38.1 \\
				\hline
				FC~\cite{rennie2017self} & 6.5 & 3.03 & 89.7 & 102.7 & 43.2 & 31.2 \\
				Att2in~\cite{rennie2017self} & 10.8 & 2.65 & 88.0 & 116.7   & 48.0 & 35.5 \\
				UpDown~\cite{updown} & 12.9 & 2.55 & 86.7 & 121.5   & 49.2 & 36.8 \\
				AoANet~\cite{huang2019attention}* & 14.6 & 2.47 & 87.2 &128.6 & 50.4 & 38.2 \\
				\hline
				DiscCap~\cite{2_disccap} & 14.0 & 2.48 & 89.2 & 120.1 & 48.5 & 36.1 \\
				CL-Cap~\cite{1_contrastive} & 14.2 & 2.54 & 81.3 & 114.2 & 46.0 & 35.3 \\
				CIDErBtwCap~\cite{17_wang2020compare} & 15.9 & 2.39 & 82.7 & 127.8  & 51.0 & 38.5 \\
				
		\end{tabular}}
		\caption{Comparison of caption distinctiveness and accuracy on MSCOCO test split: \textbf{DisWordRate}~(D), \textbf{CIDErRank}~(CR), and \textbf{CIDErBtw}~(CB) measure the distinctiveness, while \textbf{CIDEr}~(C) and \textbf{BLEU}~(B3 and B4) measure the accuracy. $\uparrow$ and $\downarrow$ show whether higher or lower score are better according to each metric. We apply our model on four baseline models: Transformer~\cite{vaswani2017attention} and M$^2$Transformer~\cite{cornia2020meshed} trained only with cross entropy loss, Transformer + SCST~\cite{vaswani2017attention} and M$^2$Transformer + SCST~\cite{cornia2020meshed} trained with reinforcement learning. $*$ denotes we train the model from scratch with official released code.  
		}
		\label{table:main_results_shrink}
	\end{center}
\end{table}

\CUT{
	\begin{table*}[tb]
		\begin{tabular}{l|ccc|ccc}
			\textbf{Method} &  \textbf{DisWordRate(\%)$\uparrow$} & \textbf{CIDErRank$\downarrow$} & \textbf{CIDErBtw$\downarrow$} &\textbf{CIDEr$\uparrow$}    & \textbf{BLEU3$\uparrow$} & \textbf{BLEU4$\uparrow$}  \\
			\hline
			M$^2$Transformer & 16.4 & 2.52 & 76.8 & \textbf{111.8}   & 45.2 & 34.7  \\
			\hspace{2mm}+ ImageGroup & 16.9 & 2.51 & 75.4  &110.7   & 45.2 & 34.8 \\
			\hspace{4mm}+ DisLoss & 17.1 & 2.48 & 76.4 & 110.0   & \textbf{45.9} & \textbf{35.5} \\
			\hspace{6mm} + GMA & 18.4 & 2.44  & 73.5 & 111.2   & 45.4 & 34.9 \\
			\hspace{8mm} + MemCls & \textbf{18.7}  & \textbf{2.42} & \textbf{72.5} &  109.8  & 45.6 & 34.7 \\
		\end{tabular}
		\caption{Ablation Study. We train M$^2$Transformer with cross-entropy loss~(XE) as the baseline, and gradually add four components of our full model: image group based training, distinctive word loss, group-based memory attention, and memory classification loss.}
		\label{table:ablation}
	\end{table*}
}

\begin{table}[tb]
	\begin{tabular}{l|ccc|ccc}
		\textbf{Method} &  \textbf{D(\%)$\uparrow$} & \textbf{CR$\downarrow$} & \textbf{CB$\downarrow$} &\textbf{C$\uparrow$}    & \textbf{B3$\uparrow$} & \textbf{B4$\uparrow$}  \\
		\hline
		M$^2$Transformer & 16.4 & 2.52 & 76.8 & \textbf{111.8}   & 45.2 & 34.7  \\
		\hspace{2mm}+ ImageGroup & 16.9 & 2.51 & 75.4  &110.7   & 45.2 & 34.8 \\
		\hspace{4mm}+ DisLoss & 17.1 & 2.48 & 76.4 & 110.0   & \textbf{45.9} & \textbf{35.5} \\
		\hspace{6mm} + GMA & 18.4 & 2.44  & 73.5 & 111.2   & 45.4 & 34.9 \\
		\hspace{8mm} + MemCls & \textbf{18.7}  & \textbf{2.42} & \textbf{72.5} &  109.8  & 45.6 & 34.7 \\
	\end{tabular}
	\caption{Ablation Study. We train M$^2$Transformer with cross-entropy loss~(XE) as the baseline, and gradually add four components of our full model: image group based training, distinctive word loss, group-based memory attention, and memory classification loss.}
	\label{table:ablation_shrink}
\end{table}

In the following, we present a comparison with the state-of-the-art (distinctive) image captioning models. In addition, we present an ablation study of applying our model to the baseline method.



\myparagraph{Comparison with the state-of-the-art.} 
We compare our model with two groups of state-of-the-art models: 1) FC~\cite{rennie2017self}, Att2in~\cite{rennie2017self}, UpDown~\cite{updown} and AoANet~\cite{huang2019attention} that aim to generate captions with high accuracy; 2) DiscCap~\cite{2_disccap}, CL-Cap~\cite{1_contrastive}, and CIDErBtwCap~\cite{17_wang2020compare} that generate distinctive captions. 

The main experiment results are presented in Table~\ref{table:main_results_shrink}, and we make the following observations.
First, when applied to four baseline models, our model achieves impressive improvement for the distinctive metrics, while maintaining comparable results on accuracy metrics. For example, we improve the DisWordRate by $16.1$ percent~(from $16.8\%$ to $19.5\%$) and reduce the CIDErBtw by $5.2$ percent~(from $74.8$ to $70.9$) for Transformer, while only sacrificing the CIDEr by $3.9$ percent. Second, \abcaddpre{in terms of distinctiveness,} models trained with cross-entropy loss tend to perform better than models trained with SCST~\cite{rennie2017self}. For example, we achieve the highest DisWordRate with Transformer + GdisCap at $19.5\%$. M$^2$Transformer + GdisCap also achieves higher DisWordRate than M$^2$Transformer + SCST + GdisCap.
Third, compared with state-of-the-art models that improve the accuracy of generated captions, our model M$^2$Transformer + SCST + GdisCap achieves comparable accuracy, while gaining impressive improvement in distinctness -- we obtain comparable CIDEr with AoANet~($127.5$ vs $128.6$), and \wenjiaadd{attain significantly higher DisWordRate, i.e., $14.6\%$~(AoANet) vs $18.5\%$~(ours).}
When compared to other models that focus on distinctiveness, M$^2$Transformer + SCST + GdisCap \wenjiaadd{achieves higher distinctiveness by a large margin -- we gain DisWordRate by $18.5\%$ compared with $15.9\%$~(CIDErBtwCap), and we obtain much lower CIDErBtw by $81.3$ vs.~$89.2$~(DiscCap).}


\myparagraph{Ablation study.} To measure the influence of each component in our \texttt{GdisCap}, we design an ablation study where we train the baseline M$^2$Transformer with cross-entropy loss.
Four variants of \texttt{GdisCap} are trained by gradually adding the components, \ie, image group based training, distinctive word loss, group-based memory attention (\texttt{GMA}) module, and memory classification loss, to the baseline model. The results are shown in Table~\ref{table:ablation_shrink}, and demonstrate that the four additional components improve the distinctive captioning metrics consistently. As pointed out in~\cite{19_wang2019describing}, increasing the distinctiveness of generated captions sacrifices the accuracy metrics such as CIDEr and BLEU, since the distinctive words cannot agree with all the GT captions due to the diversity of human language. Applying our model on top of M$^2$Transformer increases the DisWordRate by $14\%$ (from $16.4\%$ to $18.7\%$), while only sacrificing $1.8\%$ of the CIDEr value (from $111.8$ to $109.8$). More ablation studies are in the Supplementary.

\begin{figure}[tbh]
	\begin{center}
		\includegraphics[width=.95\linewidth]{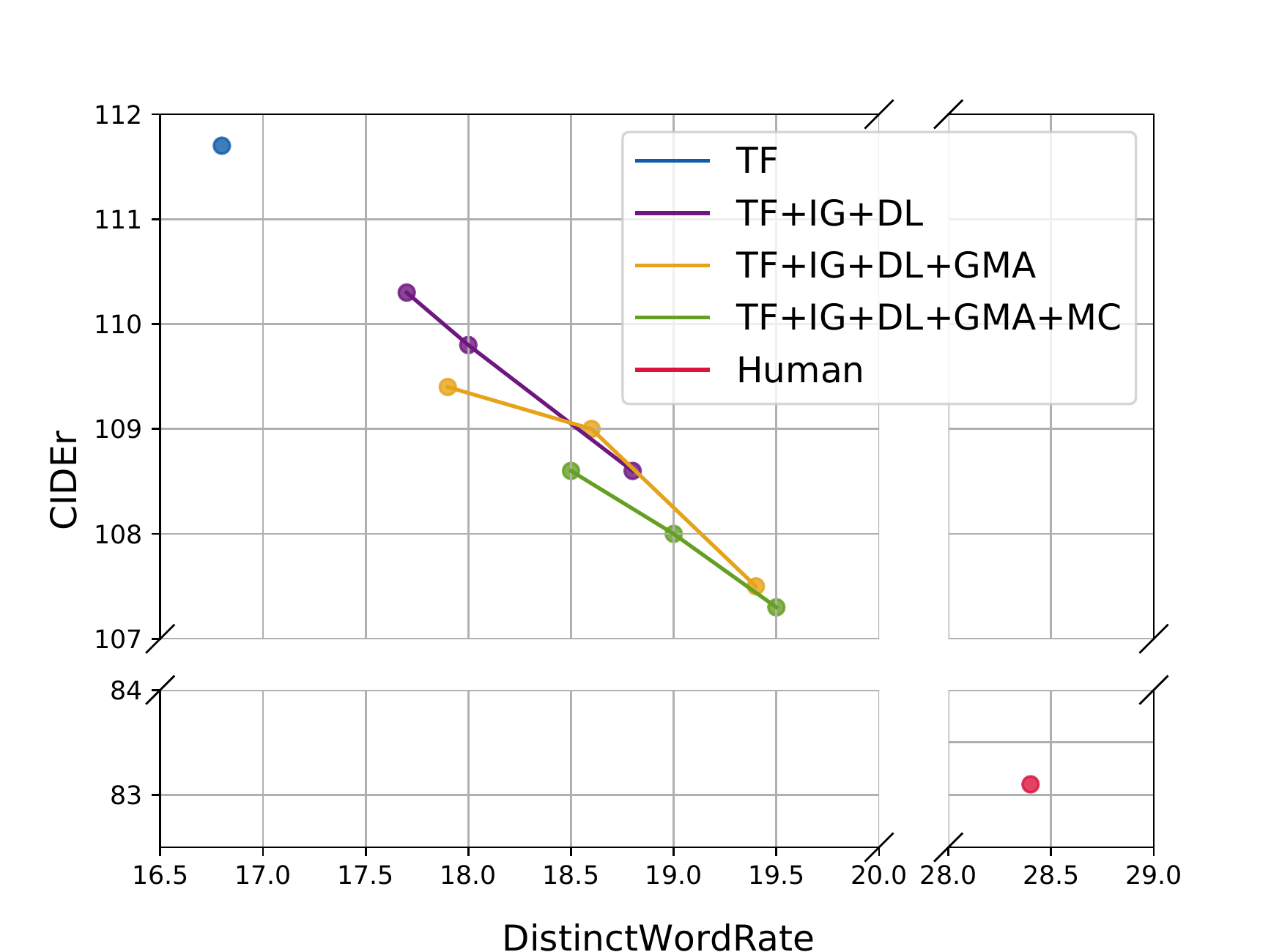}
	\end{center}
	\caption{The trade-off between accuracy (CIDEr) and distinctiveness (DisWordRate): \wenjiaadd{
			human-annotated GT captions~(Human), baseline model Transformer~(TF)~\cite{vaswani2017attention}, and three variants of our model using different components.
			IG, DL, MA and MC denote ImageGroup, DisLoss, Group-based Memory Attention and MemCls. For our models, we show three training stages (at different epochs), 
			which demonstrates 
			the trade-off between accuracy and distinctiveness during training.} 
	}
	\label{fig:trade_off}
\end{figure}

\myparagraph{Trade-off between accuracy and distinctiveness.}
\wenjiaadd{The results in Figure~\ref{fig:trade_off} demonstrate that improving distinctiveness typically hurts the accuracy, since the distinctive words do not appear in all the GT captions, while CIDEr considers the overlap between the generated captions and all GT captions. This can explain why the human-annotated GT captions, which are considered the upper bound of all models, only achieve CIDEr of $83.1$. Compared to the baselines, our work achieves results more similar to human performance.}


\section{User Study}
To fairly evaluate the distinctiveness of our model from the human perspective, we propose a caption-image retrieval user study, which extends the evaluation protocol proposed in ~\cite{17_wang2020compare,2_disccap}. Each test is a tuple $\left( {I_0, I_1, \dots, I_K}, \hat{c} \right)$, which includes a similar image group
and a caption generated by a random model describing a random image in the group. The users are instructed to choose the target image that the caption corresponds to. To evaluate one image captioning model, we randomly select 50 tuples with twenty participants for each test. A correct answer is regarded as a hit, and the accuracy scores for twenty participants are averaged to obtain the final retrieval accuracy. A higher retrieval score indicates more distinctiveness of the generated caption, \ie, it can distinguish the target image from other similar images (more details are in the Supplementary).

\wenjiaadd{We compare our \texttt{GdisCap} model with three competitive models, DiscCap~\cite{2_disccap}, CIDErBtwCap~\cite{17_wang2020compare}, and M$^2$Transformer + SCST~\cite{cornia2020meshed}. The results are shown in Table~\ref{table:user_study}, where our model achieves the highest caption-image retrieval accuracy -- $68.2$ compared to M$^2$Transformer + SCST with $61.9$. The user study demonstrates that our model generates the most distinctive captions that can distinguish the target image from the other images with similar semantics. The results agree with the DisWordRate and the CIDErRank displayed in Table~\ref{table:main_results_shrink}, which indicates that the proposed two metrics are effective evaluations similar to human judgment.}


\begin{table}[tb]
\begin{center}
\small
\begin{tabular}{@{}c|cccc@{}}
Method & DiscCap \cite{2_disccap} & CIDErBtwCap \cite{17_wang2020compare} & M$^2$+SCST \cite{cornia2020meshed} & \emph{Ours} \\
\hline
Accuracy & 48.1 & 58.7 & 61.9 & {\bf 68.2} \\
\end{tabular}

\caption{User study results for caption-image retrieval. 
Our model produces captions with significantly higher retrieval accuracy (2-sample z-test on proportions, $p<0.01$).}
\label{table:user_study}
\end{center}
\end{table}


\begin{figure*}[t!]
	\begin{center}
		\includegraphics[width=\linewidth]{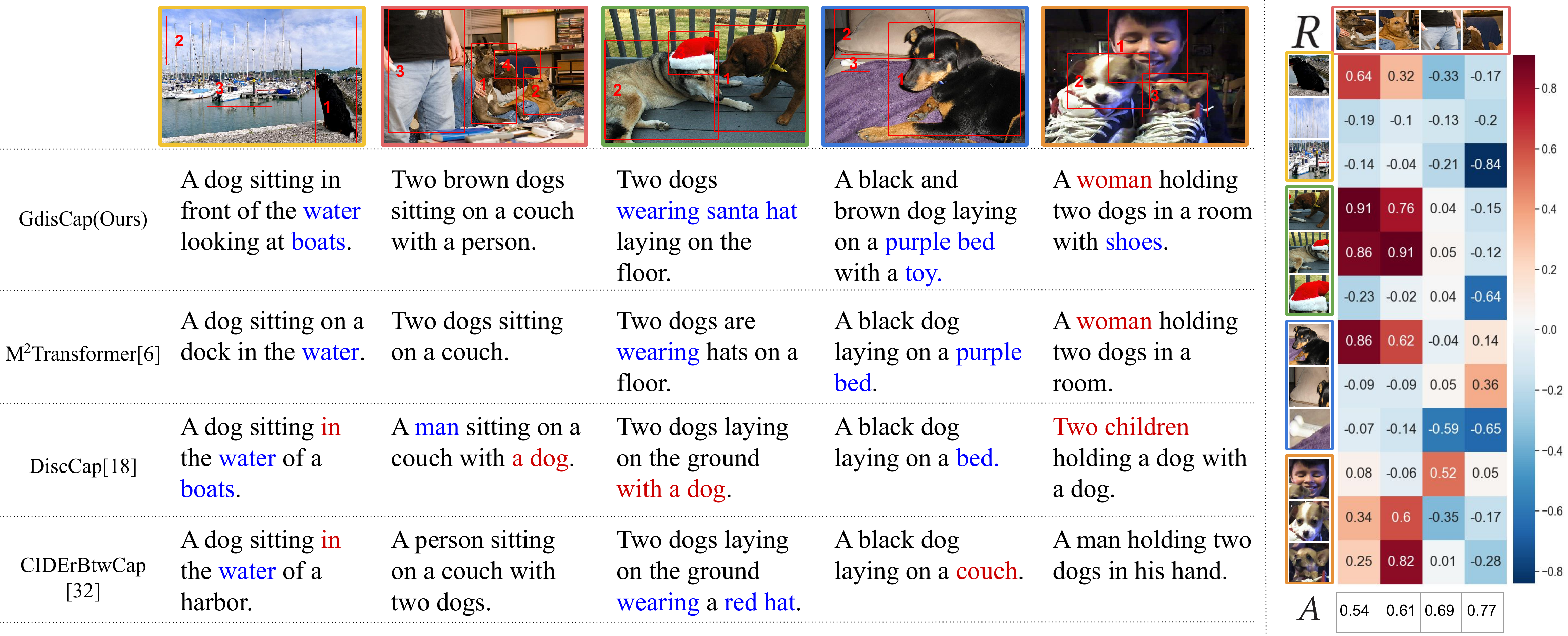}
	\end{center}
	\caption{Qualitative results. Left: Captions for one similar image group with five images from the test set. We compare our model with three state-of-the-art methods, M$^2$Transformer~\cite{cornia2020meshed}, DiscCap~\cite{2_disccap}, and CIDErBtwCap~\cite{17_wang2020compare}. The blue words indicate the distinctive words $w_{d}$ that appear in GT captions of the target image, while not in captions of similar images. The red words denote the mistakes in the generated captions. Right: Visualization of the similarity matrix $R$ and distinctive attention $A$. Here we show the second image~(in red box) as the target $I_0$ and display the similarity value between four salient objects in $I_0$ and the objects in the other four images. \abcaddd{The attention $A$ denotes the overall distinctiveness of each object in the image $I_0$.} 
		Objects in the same colored box are from the same image. 
	}
	\label{fig:quali_2}
\end{figure*}

\section{Qualitative Results}
\wenjiaadd{In this section, we first qualitatively evaluate our model in generating distinctive captions for similar image groups. Second, we visualize the group-based memory attention calculated by our model to highlight the distinct objects.}
%
We compare our \texttt{GdisCap} with  three other models, M$^2$Transformer~\cite{cornia2020meshed}, DiscCap~\cite{2_disccap}, and CIDErBtwCap~\cite{17_wang2020compare}. which are the best  competing methods on distinctiveness. \abcaddd{More examples are found in the supplemental.}

Figure~\ref{fig:quali_2}~(left) displays the captions generated by four models for one similar image group. Overall, all the methods generate accurate captions specifying the salient content in the image. However, their performances on the distinctiveness differ.
M$^2$Transformer and DiscCap generate captions that only mention the most salient objects in the image, using less distinctive words. 
For instance, in Figure~\ref{fig:quali_2}~(column 1), our \texttt{GdisCap} generates captions ``a dog sitting in front of the water looking at boats'', compared to the simpler caption  ``a dog sitting on a dock in the water'' from M$^2$Transformer. Similarly, in Figure~\ref{fig:quali_2}~(column 3), our \texttt{GdisCap} describes the most distinctive property of the target image, the ``santa hat'', compared to DiscCap that only provides  ``two dogs laying on the ground''.
The lack of distinctiveness from M$^2$Transformer and DiscCap is due to 
the models being supervised by equally weighted GT captions, which tends to produce generic words that agree with all the supervisory captions. 

CIDErBtwCap, on the other hand, reweights the GT captions according to their distinctiveness, and thus generates captions with more distinctive words. Compared to CIDErBtwCap, where all the objects in the image are attached with the same attention, our method yields more distinctive captions that  distinguish the target image from others by attaching higher attention value to the unique details and objects that appear in the image. 
For example, in Figure~\ref{fig:quali_2}~(column 3), \texttt{GdisCap} describes the distinctive ``santa hat'', while  CIDErBtwCap mentions it as a ``red hat''. 

Remarkably, \texttt{GdisCap} is more aware of the locations of objects in the image and the relationships among them.
For example, in Figure~\ref{fig:quali_2}~(column 5), our caption specifies the ``A woman holding two dogs in a room with shoes'', compared with CIDErBtwCap, which wrongly describe ``holding two dogs in his hand'' when no hands appear on the image. 
It is interesting because there is no location supervision for different objects, but our model learns the relation solely from the GT captions. \abcaddd{Finally, Figure~\ref{fig:quali_2}~(right) displays the similarity matrix $R$ and distinctive attention $A$ for the 2nd image as the target image.  The object regions with highest attention are those with lower similarity to the objects in other images, in this case the ``couch'' and the ``person''.  The ``dogs'', which are the common objects among the images, have lower non-zero attention so that they are still described in the caption.}

\section{Conclusion}

\wqz{In this paper, we have investigated a vital property of image captions -- distinctiveness, which mimics the human ability to describe the unique details of images, so that the caption can distinguish the image from other semantically similar images. We presented a Group-based Distinctive Captioning Model (\texttt{GdisCap}) that compares the objects in the target image to objects in semantically similar images and highlights the unique image regions. Moreover, we developed two loss functions to train the proposed model: the distinctive word loss encourages the model to generate distinguishing information; the memory classification loss helps the weighted memory attention to contain distinct concepts. We conducted extensive experiments and evaluated the proposed model using multiple metrics, showing that the proposed model outperforms its counterparts quantitatively and qualitatively. Finally, our user study verifies that our model indeed generates distinctive captions based 
	human judgment.}

\clearpage

\bibliographystyle{ACM-Reference-Format}
\balance
\bibliography{egbib}

\end{document}


\fancyhead[R]{\small Jiuniu Wang, Wenjia Xu, Qingzhong Wang, Antoni B. Chan}
\fancyhead[L]{\small Group-based Distinctive Image Captioning with Memory Attention}

\title{Group-based Distinctive Image Captioning with \\ Memory Attention}


\author{Jiuniu Wang$^{1,2,3}$, Wenjia Xu$^{2,3}$, Qingzhong Wang$^{1,4}$, Antoni B. Chan$^1$} 

\affiliation{\institution{$^1$ Department of Computer Science, City University of Hong Kong}}
\affiliation{\institution{$^{2}$ Aerospace Information Research Institute, Chinese Academy of Sciences}}
\affiliation{\institution{$^{3}$ University of Chinese Academy of Sciences}}
\affiliation{\institution{$^{4}$ Baidu Research}}
\affiliation{\institution{\{jiuniwang2-c,qingzwang2-c\}@my.cityu.edu.hk, xuwenjia16@mails.ucas.ac.cn, abchan@cityu.edu.hk}}
\renewcommand{\shortauthors}{Jiuniu Wang, Wenjia Xu, Qingzhong Wang, Antoni B. Chan}
\appendix
\counterwithin{figure}{section}
\counterwithin{table}{section}
\begin{center}
{\textbf{\huge Supplementary Material for Group-based Distinctive Image Captioning \\ with Memory Attention}}
\end{center}
\vspace{20pt}

The organization of the supplementary material is as follows. We include the interface for the caption-image retrieval user study in Section~\ref{section:sp_user_study}, then we show more ablation study results in Section~\ref{section:Ablation_Study} and more qualitative analysis in Section~\ref{section:Qualitative_results}.
\section{User Study}

\begin{figure*}[h]
	\begin{center}
		\includegraphics[width=.8\linewidth]{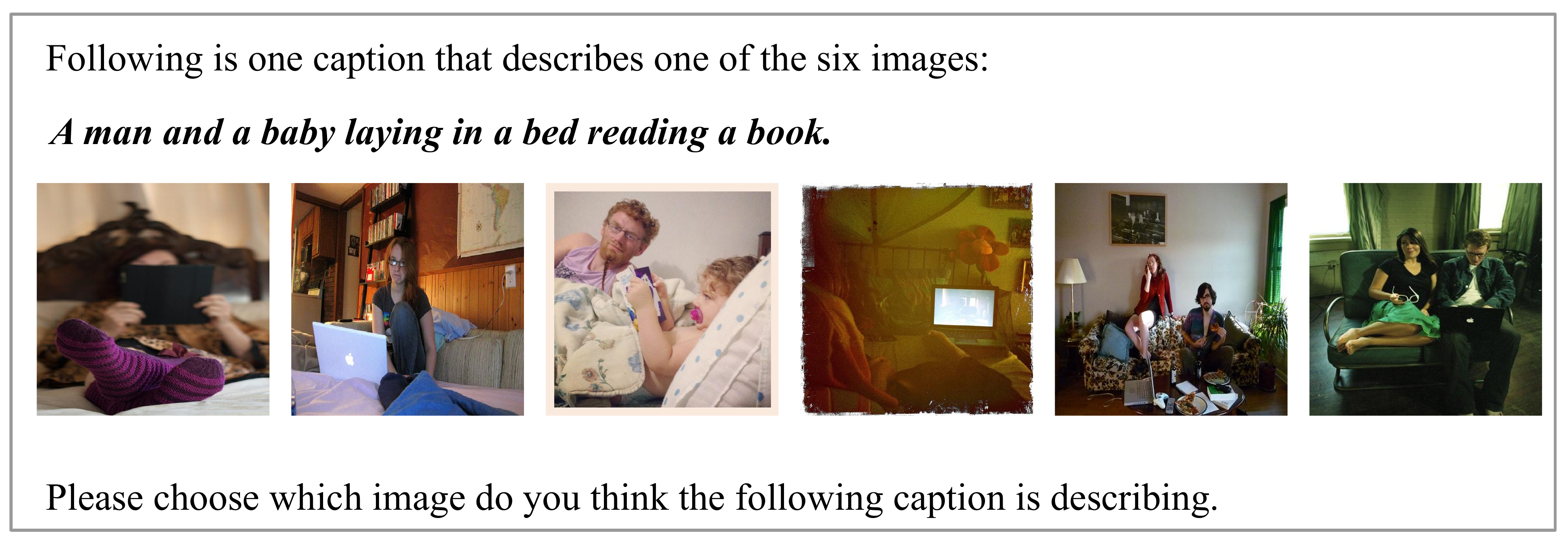}
	\end{center}
	\caption{The user study interface. We display a group of six similar images, and a caption generated from one image by an image captioning model. The users are asked to select the image that they think the caption is describing. }
	\label{fig:user_study}
\end{figure*}

In Figure~\ref{fig:user_study}, we display the interface for our user study.
\label{section:sp_user_study}

\section{Ablation Study}
\label{section:Ablation_Study}
We show more results of the ablation study in Table~\ref{table:ablation_2}, where we train M$^2$Transformer + SCST~\cite{cornia2020meshed} as the baseline and add four components of our model gradually. The results demonstrate that the four additional components improve the distinctive captioning metrics consistently.

\begin{table*}[h]
\begin{tabular}{l|ccc|ccc}
\textbf{Method} &  \textbf{DisWordRate(\%)$\uparrow$} & \textbf{CIDErRank$\downarrow$} & \textbf{CIDErBtw$\downarrow$} &\textbf{CIDEr$\uparrow$}    & \textbf{BLEU3$\uparrow$} & \textbf{BLEU4$\uparrow$}  \\
\hline
 
 M$^2$Transformer + SCST & 17.3 & 2.38 & 82.9 & \textbf{128.9}    & \textbf{50.6} & \textbf{38.7}  \\
  \hspace{2mm}+ ImageGroup & 17.4 & 2.35 & 82.3 & 127.8   & 50.4 & 38.5  \\
  \hspace{4mm} + DisLoss & 18.2 & 2.32 & 82.0 & 128.4   & 50.4 & 38.5 \\
  \hspace{6mm} + GMA & 18.3 & 2.34 & 81.1 &  127.8   & 50.5 & 38.6  \\
  \hspace{8mm} + MemCls & \textbf{18.5} & \textbf{2.31} & \textbf{81.3} &127.5 & 50.0 & 38.1
\end{tabular}
\caption{Ablation Study. We train M$^2$Transformer + SCST as the baseline model, and gradually add the four components of our full model: image group based training, distinctive word loss, group-based memory attention, and memory classification loss.}
\label{table:ablation_2}
\end{table*}

\section{More Qualitative Results}
\label{section:Qualitative_results}
\begin{figure*}[tb]
	\begin{center}
		\includegraphics[width=\linewidth]{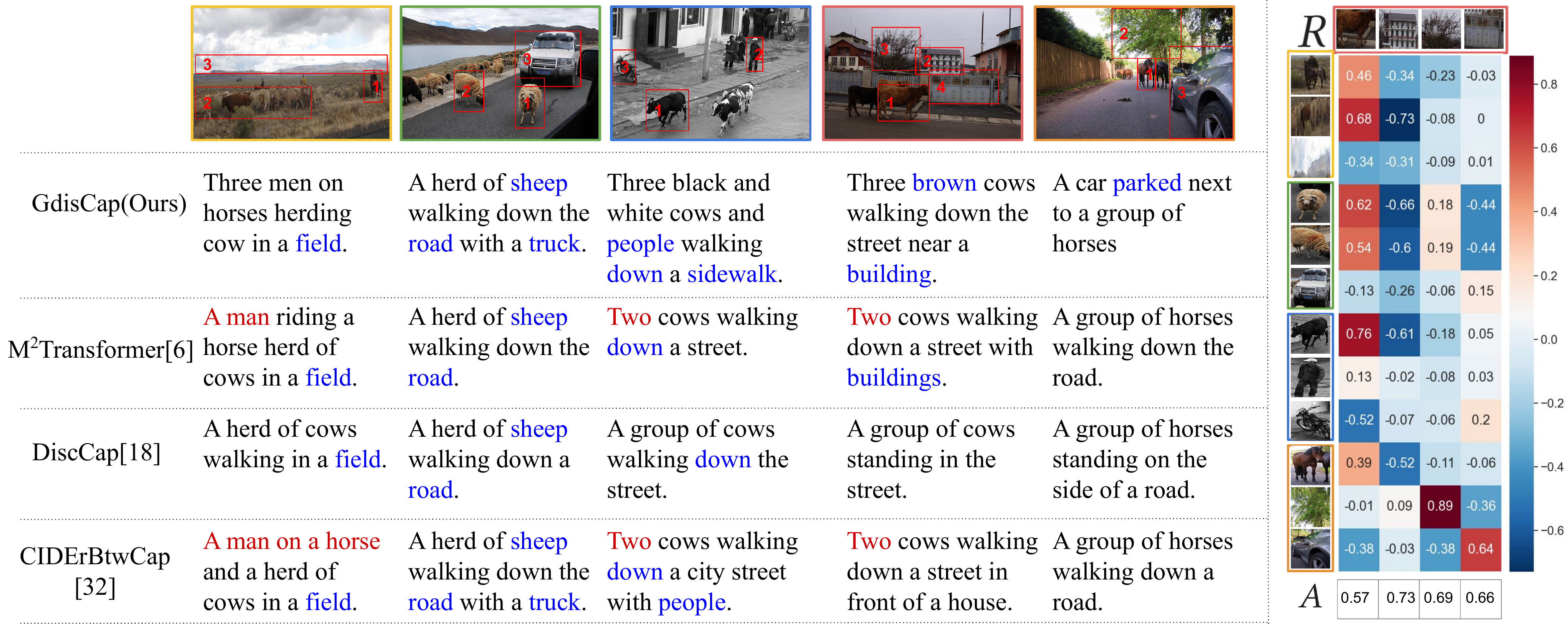}
	\end{center}
	\caption{Qualitative results. Left: Captions for one similar image group with five images from the test set. The blue words indicate the distinctive words $w_{d}$ that appear in GT captions of the target image, while not in captions of similar images. The red words denote the mistakes in the generated captions. Right: Visualization of the similarity matrix $R$ and distinctive attention $A$. Here we show the fourth image~(in red box) as the target $I_0$ and display the similarity value between four salient objects in $I_0$ and the objects in the other four images. \abcaddd{The attention $A$ denotes the overall distinctiveness of each object in the image $I_0$.} Objects in the same colored box are from the same image. }
	\label{fig:quali_1}
\end{figure*}

\begin{figure*}[tb]
	\begin{center}
		\includegraphics[width=\linewidth]{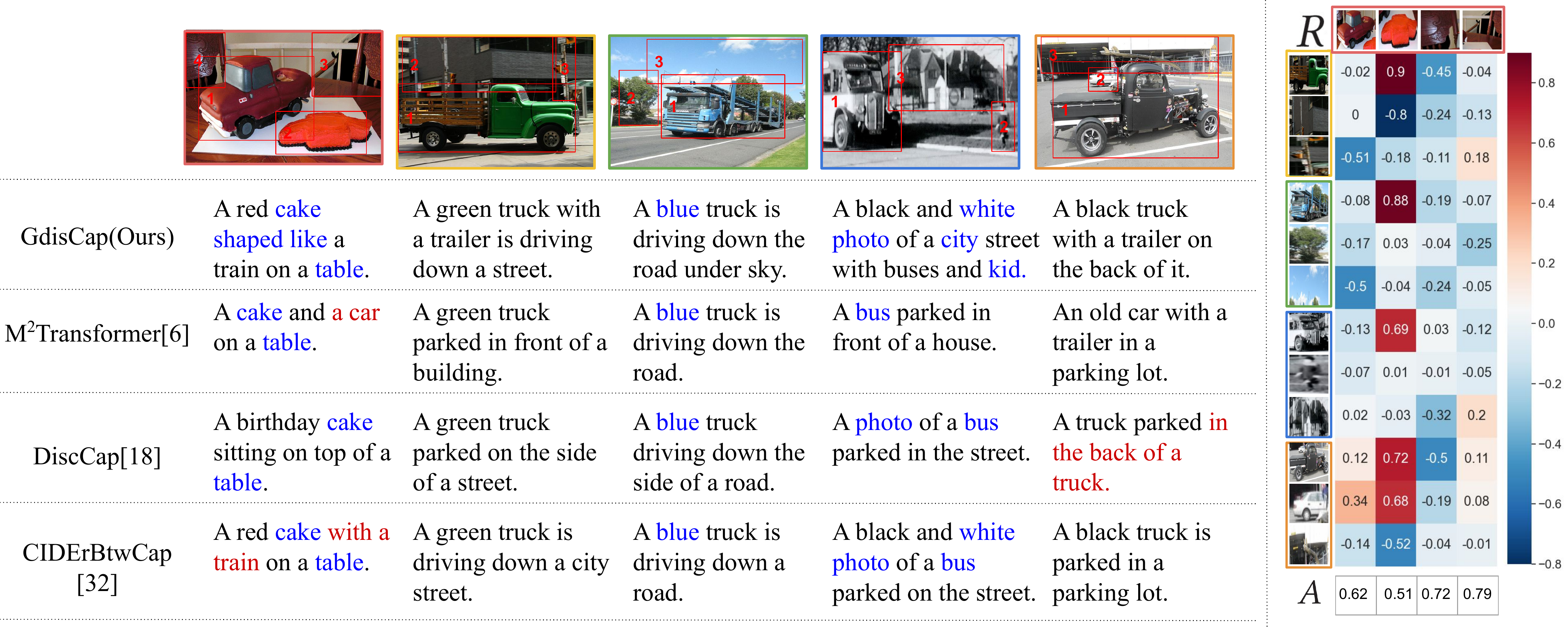}
	\end{center}
	\caption{Qualitative results. Left: Captions for one similar image group with five images from the test set. The blue words indicate the distinctive words $w_{d}$ that appear in GT captions of the target image, while not in captions of similar images. The red words denote the mistakes in the generated captions. Right: Visualization of the similarity matrix $R$ and distinctive attention $A$. Here we show the first image~(in red box) as the target $I_0$ and display the similarity value between four salient objects in $I_0$ and the objects in the other four images. \abcaddd{The attention $A$ denotes the overall distinctiveness of each object in the image $I_0$.} Objects in the same colored box are from the same image.}
	\label{fig:quali_3}
\end{figure*}

\wenjiaadd{We show more qualitative results in Figure~\ref{fig:quali_1} and Figure~\ref{fig:quali_3}.
Remarkably, our model is better at counting. For example, in Figure~\ref{fig:quali_1}~(columns 3 and 4), \texttt{GdisCap} correctly describes the count and the color of the cows, while all other methods do not provide these details. Another example is observed in Figure~\ref{fig:quali_1}~(column 1), where the caption ``three man on horses herding cow’’ captures the small but salient objects correctly. This is due to the effect of weighted memory attention, where each distinct object memory is tied with high attention. In addition, our GdisCap is good at describing details such as color and small objects that are not observed by other models, e.g., in Figure~\ref{fig:quali_3}~(columns 2) our model mentions the kid in front of the bus; in Figure~\ref{fig:quali_3}~(columns 1), we correctly describe "A red cake shaped like a train on a table", compared to M$^2$Transformer and CIDErBtwCap that wrongly say "A cake and a train on the table". We display the similarity matrix $R$ and distinctive attention $A$ on the right side.  The object regions with the highest attention are those with lower similarity to the objects in other images, i.e., the building in Figure~\ref{fig:quali_1}, and the table in Figure~\ref{fig:quali_3}. }

\section{Algorithm}
We show the details for processing one image group in Algorithm~\ref{algo:1}. During training, each mini-batch comprises several similar image groups, with the loss aggregated over each image as a target in its group.

\renewcommand{\algorithmicrequire}{\textbf{Input:}}
\renewcommand{\algorithmicensure}{\textbf{Output:}}
\begin{algorithm} 
	\caption{The training procedure of \texttt{GdisCap} in each step} 
	\label{alg} 
	\begin{algorithmic}[1] 
		\Require A similar image group ${I_0,\dots, I_K}$ with captions ${C_0,…,C_{K}}$ 
		\Ensure The final loss $ \mathcal{L}$ of this similar image group to optimize the {\em Encoder} and {\em Decoder} 
		
		\State Encode the image group $\{ {I_0,\dots, I_K} \}$ into $\{M_0,\dots,M_K\}$, where the memory of  the $k$-th image is $M_k = \{m_k^i\}_{i=1}^{N_k}$
		\For{$k \gets 0$ to $K$} 
		\State Calculate the distinctive attention $A = f_{GMA}(M_k)$, where $A=\{ a_i \}_{ i=1}^{N_k} \in R^{N_k}$
		\State Weight $M_k$ into $M_k’$: $M_k’=\{ a_i \cdot m_k^i\}_{i=1}^{N_k}$
		\State Calculate distinctive word $w_{d} = f_{set}(C_k) - f_{set}(\{C_j\}_{j\neq k})$
		\State Decode $M_k’$ as probability of generated words $\{P_t\}_{t=1}^T$
		\State Classify $M_k’$ as possible words $P_M \gets f_{MC}(M_k')$
		\State Calculate each loss for the $k$-th image, including: 
		\State \quad Cross-entropy loss $\mathcal{L}_{xe}$ by $P_t$ and $C_k$, 
		\State \quad Reinforcement learning loss $\mathcal{L}_c$ by $P_t$ and $C_k$,
		\State \quad Distinctive word loss $ \mathcal{L}_d$ by $P_t$ and $w_{d}$, 
		\State \quad Memory classification loss $\mathcal{L}_m$ by $P_M$ and $w_{d}$
		\State Get the loss for the $k$-th image $\mathcal{L}_k = \alpha_{c} \mathcal{L}_{xe} +  \alpha_{r} \mathcal{L}_r +  \alpha_{d} \mathcal{L}_d +  \alpha_{m} \mathcal{L}_m$
		\EndFor
		\State Accumulate the loss $\mathcal{L} = \sum\limits_{k=0}^K \mathcal{L}_k $
		
	\end{algorithmic}
	\label{algo:1}
\end{algorithm}

\bibliographystyle{ACM-Reference-Format}
\balance
\bibliography{egbib}
\clearpage
\onecolumn